\def\year{2022}\relax
\documentclass[letterpaper]{article} % DO NOT CHANGE THIS
\usepackage{aaai22}  % DO NOT CHANGE THIS
\usepackage{times}  % DO NOT CHANGE THIS
\usepackage{helvet} % DO NOT CHANGE THIS
\usepackage{courier}  % DO NOT CHANGE THIS
\usepackage[hyphens]{url}  % DO NOT CHANGE THIS
\usepackage{graphicx} % DO NOT CHANGE THIS
\usepackage{natbib}  % DO NOT CHANGE THIS AND DO NOT ADD ANY OPTIONS TO IT
\usepackage{caption} % DO NOT CHANGE THIS AND DO NOT ADD ANY OPTIONS TO IT
\usepackage{times}
\usepackage{epsfig}
\usepackage{graphicx}
\usepackage{amsmath}
\usepackage{amssymb}
\usepackage{caption}
\usepackage{tabularx}
\usepackage{booktabs}       % professional-quality tables
\usepackage{amsfonts}       % blackboard math symbols
\usepackage{nicefrac}       % compact symbols for 1/2, etc.
\usepackage{microtype}      % microtypography
\usepackage{mathtools}
\usepackage{arydshln}
\usepackage{xspace}
\usepackage{bm}

\usepackage{xcolor}

\title{Multi-Modal Answer Validation for Knowledge-Based VQA}

\author {
     Jialin Wu\,\textsuperscript{\rm 1},
     Jiasen Lu\,\textsuperscript{\rm 2},
     Ashish Sabharwal\,\textsuperscript{\rm 2},
     Roozbeh Mottaghi\,\textsuperscript{\rm 2} \\
 }
 \affiliations {
     \textsuperscript{\rm 1} The University of Texas at Austin \\
     \textsuperscript{\rm 2} Allen Institute for AI \\
     jialinwu@utexas.edu, \{jiasenl, ashishs, roozbehm\}@allenai.org

 }

\newcommand{\primemacro}[1]{#1^\prime}
\newcommand{\mavex}{{MAVEx}\xspace}

\begin{document}

\maketitle

\begin{abstract}
The problem of knowledge-based visual question answering involves answering questions that require external knowledge in addition to the content of the image. Such knowledge typically comes in various forms, including visual, textual, and commonsense knowledge. Using more knowledge sources increases the chance of retrieving more irrelevant or noisy facts, making it challenging to comprehend the facts and find the answer. To address this challenge, we propose \textbf{M}ulti-modal \textbf{A}nswer \textbf{V}alidation using \textbf{Ex}ternal knowledge (\mavex), where the idea is to validate a set of promising answer candidates based on answer-specific knowledge retrieval. Instead of searching for the answer in a vast collection of often irrelevant facts as most existing approaches do, \mavex aims to learn how to extract relevant knowledge from noisy sources, which knowledge source to trust for each answer candidate, and how to validate the candidate using that source.
Our multi-modal setting is the first to leverage external visual knowledge (images searched using Google), in addition to textual knowledge in the form of Wikipedia sentences and ConceptNet concepts. Our experiments with OK-VQA, a challenging knowledge-based VQA dataset, demonstrate that \mavex achieves new state-of-the-art results. Our code is available at \url{https://github.com/jialinwu17/MAVEX}
 \end{abstract}

\section{Introduction}
Over the past few years, the domain of Visual Question Answering (VQA) has witnessed significant progress \cite{antol2015vqa,zhu2016visual7w,hudson2019gqa}. However, there is a recent trend towards knowledge-based VQA \cite{wang2015explicit,wang2018fvqa,marino2019ok} which requires information beyond the content of the images. Besides visual recognition, the model needs to perform logical reasoning and incorporate external knowledge about the world to answer these challenging questions correctly. These knowledge facts can be obtained from various sources, such as image search engines, encyclopedia articles, and knowledge bases about common concepts and their relations. 

\begin{figure}[t]
    \centering
    \includegraphics[width=\columnwidth]{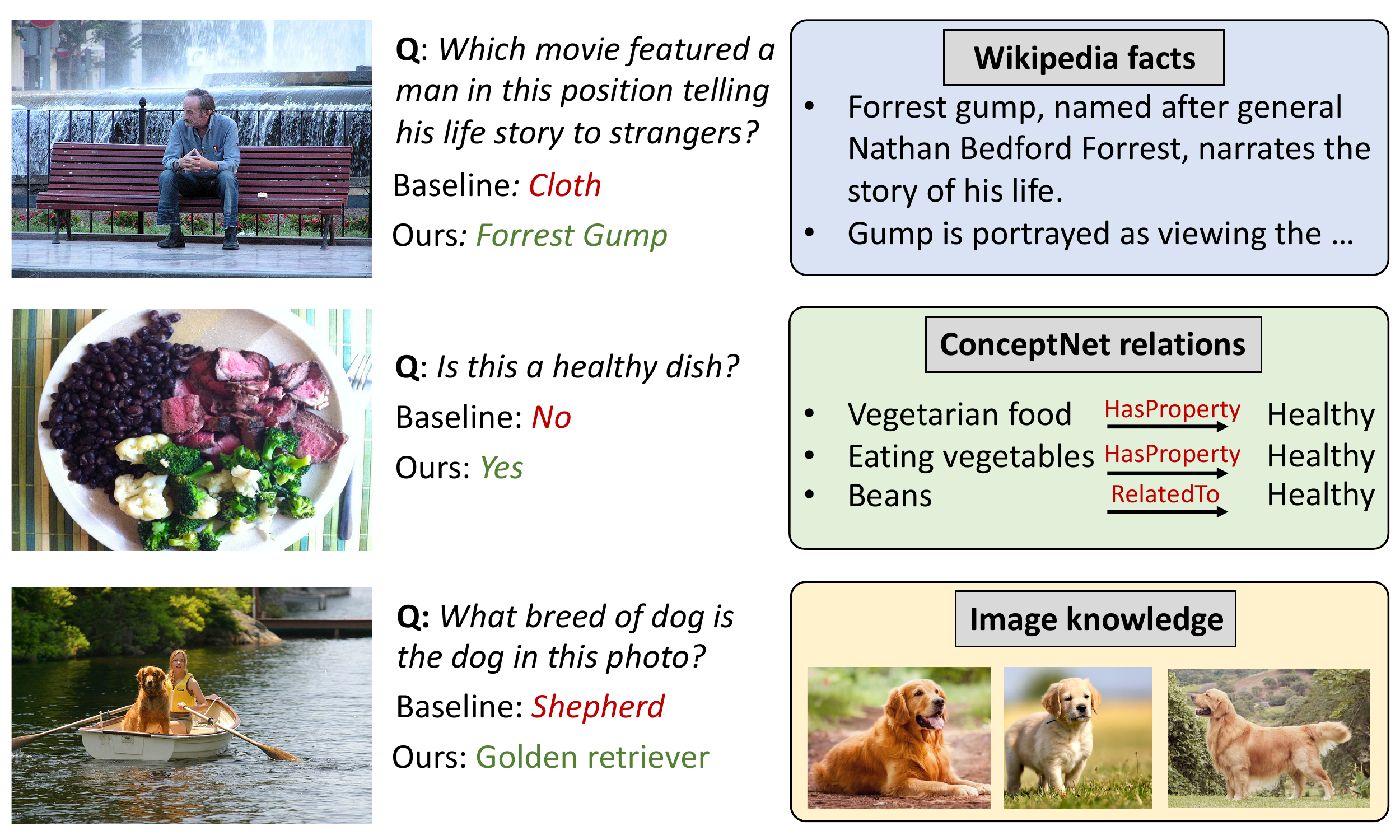}
    \caption{We address the problem of knowledge-based question answering. Retrieving relevant knowledge among diverse knowledge sources (visual knowledge, textual facts, concepts, etc.) is quite challenging. This paper aims to learn what knowledge source should be used for a particular question and how to validate a set of potential answer candidates using different knowledge sources.}
    \label{fig:teaser}
\end{figure}

Figure~\ref{fig:teaser} illustrates a few visual questions and the knowledge from different external sources that helps answer them. Each question needs a different type of external knowledge. For example, to identify the movie that featured a man telling his life story to strangers, we need to link the image content and question to some textual facts; Vegetarian food and eating vegetables are related to the concept of health; the retrieved images for a ``golden retriever.'' 
are visually similar to the dog in the question image. \emph{The challenge is to retrieve and correctly incorporate such external knowledge effectively} in an open domain question answering framework.

\begin{figure*}
    \centering
    \includegraphics[width=0.95\linewidth]{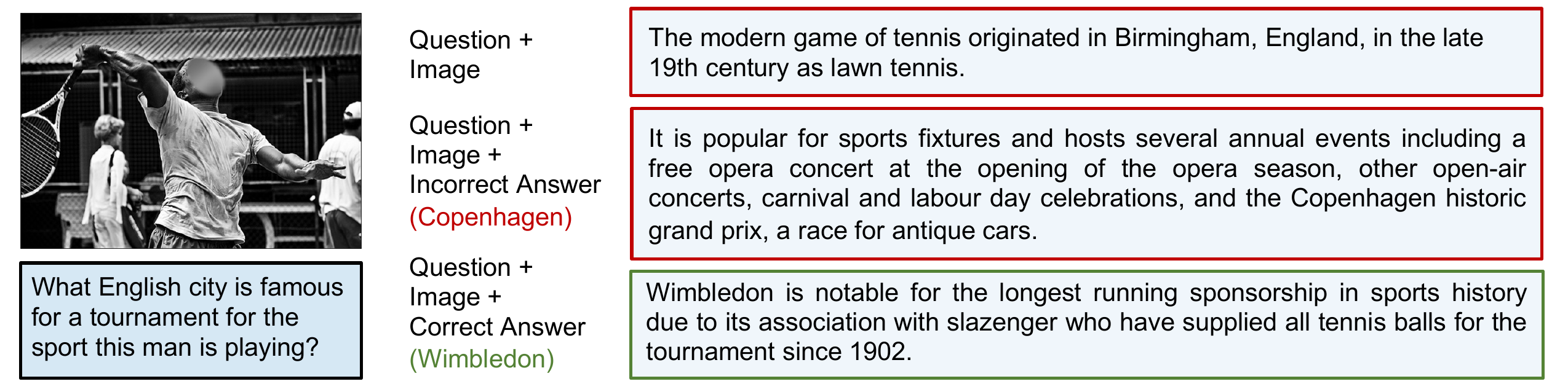}
    \caption{Examples of retrieved Wikipedia sentences using different sets of search words. The sentences retrieved using only the words in questions and objects in images (top) and the wrong answer (middle) are hardly helpful to answer the question. However, with the correct answer ``Wimbledon'' (bottom), the quality of the retrieved fact is significantly improved.}
    \vspace{-0.3cm}
    \label{fig:f1}
\end{figure*}

We also witness a shift in knowledge-based VQA datasets from structured retrieved knowledge such as triplets and dense captions \cite{wang2015explicit, wang2018fvqa} to unstructured open knowledge \cite{marino2019ok}. Most current knowledge-based VQA systems \cite{marino2019ok,wang2018fvqa,zhu2020mucko,marino2020krisp} follow a two-stage framework, where a retriever first looks up knowledge relevant to the question and the image, and then a separate comprehension model predicts the answer.

However, knowledge retrieved directly for the question and image is often noisy and not helpful in predicting the correct answer. For example, as shown in Figure~\ref{fig:f1}, the sentences retrieved using only the words in questions and objects in images (top) or a wrong answer (middle) are hardly helpful to answer the question. This increases the burden on the answer predictor, leading to only marginal improvements from the use of retrieved knowledge \cite{marino2019ok}. Interestingly, with the correct answer ``Wimbledon'' (bottom), the quality of the retrieved fact is significantly improved, making it suitable for answering the question. This observation motivates us to use retrieved knowledge for \emph{answer validation} rather than for producing the answer.

To address this challenge, we propose a new system called \mavex or \textbf{M}ulti-modal \textbf{A}nswer \textbf{V}alidation using \textbf{Ex}ternal knowledge. We use a three-stage framework. First, since state-of-the-art VQA models are surprisingly effective at generating a small set of promising answer candidates, we employ them for this purpose. Second, in the knowledge retrieval stage, we parse noun phrases from the question and answers, and generate queries for each phrase to query external resources. In order to better comprehend the retrieved knowledge, we embed it at multiple levels of granularity, from the basic query-level embedding to a noun-phrase-level embedding and finally to a question-level knowledge embedding. The goal of this multi-granular representation is to put more emphasis on queries that are important for each phrase, and on noun phrases that are critical for the question. Finally, in the validation stage, we predict how trustworthy each knowledge source is for the given question and answer candidate, and score the candidates accordingly.

We evaluate \mavex on OK-VQA~\cite{marino2019ok}, the largest knowledge-based VQA dataset to date. Our approach achieves state-of-the-art results (score 40.3, ensemble score 41.4), demonstrating that answer-specific knowledge retrieval results in more informative supporting evidence and a more solid knowledge-based VQA system.

Our main contributions are: (a) We introduce a novel approach that uses answer candidates to guide knowledge retrieval among noisy facts; (b) We leverage multi-modal knowledge by retrieving from both visual and textual resources; and (c) We demonstrate that incorporating retrieved knowledge at multiple levels of granularity, based on the question and candidate answers, is an effective strategy.  

\section{Related Work}

\noindent \textbf{Visual Question Answering.} 
Visual Question Answering (VQA) has made significant progress over the past few years \cite{antol2015vqa,lu2016hierarchical, anderson2017bottom,kim2018bilinear,ben2017mutan,cadene2019murel}. More recent VQA systems \cite{lu2019vilbert,tan2019lxmert,liu2019learning,li2019visualbert,zhou2019deep,li2020unicoder,zhou2020unified,chen2019uniter,Lu_2020_CVPR} first extract visual features from a pre-trained object detector. Then they feed both visual and textual embeddings into a multi-modal transformer, which is pre-trained on auxiliary tasks using large-scale multi-modal datasets such as \cite{sharma2018conceptual,hudson2019gqa,refcoco}. These models achieve remarkable performance on the VQA \cite{antol2015vqa} dataset; however, they can only reason based on the image content and do not have a mechanism to incorporate knowledge from external sources explicitly.  

\noindent \textbf{Knowledge-Based VQA Datasets.}  KB-VQA dataset \cite{wang2015explicit} includes 2,402 questions generated by templates for $700$ images. F-VQA \cite{wang2018fvqa} contains 5,826 questions, where each question-answer sample is annotated with a ground-truth fact triplet retrieved from the knowledge base. OK-VQA dataset \cite{marino2019ok} is a more recent open-domain dataset that covers a wide range of topics and includes 14,055 questions on 14,031 images. Our focus is on the OK-VQA dataset since it provides a larger scale dataset that requires open-domain knowledge. Due to the difficulty of collecting such datasets, knowledge-based VQA datasets are typically small compared to the traditional VQA datasets. The small scale of the datasets adds to the challenges of learning robust models.

\noindent \textbf{Knowledge-Based VQA Models.} Recent methods for knowledge-based VQA mainly follow two trends, template fitting, and learning-based approaches. For example, \cite{wang2015explicit} fit the query to several pre-defined query templates and explicitly reason about the answer using the templates. The main limitation of the template fitting approaches is that the template is hand-designed, and it is hard to accommodate the rich knowledge required to answer the questions. 

Learning-based approaches are proposed to fetch helpful facts and commonsense knowledge for better performance. \citeauthor{narasimhan2018straight} (\citeyear{narasimhan2018straight}) propose to retrieve relevant facts from a knowledge base. \citeauthor{wang2018fvqa} (\citeyear{wang2018fvqa}) desige a system to find the mappings from the question to a query triplet. GCN \cite{tompson2014joint} is applied on the fact graph in \cite{narasimhan2018out} where each node is a representation of an image-question-entity triplet. \citeauthor{li2020boosting} (\citeyear{li2020boosting}) introduce a knowledge graph augmentation approach to retrieve context-aware knowledge subgraphs and then learn to aggregate the useful visual and question-relevant knowledge. \citeauthor{zhu2020mucko} (\citeyear{zhu2020mucko}) propose a modality-aware heterogeneous GCN capturing the most supporting evidence. 

Most recent KB-VQA systems \cite{garderes2020conceptbert,marino2020krisp,shevchenko2021reasoning} utilize multi-modal transformers \cite{lu2019vilbert,li2019visualbert} as base systems to incorporate the implicit knowledge they gathered from the large scale pre-training. In particular, \cite{garderes2020conceptbert,marino2020krisp} combine the implicit knowledge with external symbolic knowledge and \cite{shevchenko2021reasoning} focus on injecting knowledge from the knowledge base into finetuning transformers. In contrast to these approaches, we formulate our problem as \emph{answer validation}, where the idea is to learn to validate a set of potential answers using multi-modal noisy knowledge sources. 

\noindent \textbf{External Knowledge in Knowledge-Based VQA.} To answer knowledge-based visual questions, most systems acquire external knowledge from various textual resources. For example, Wikipedia articles and ConceptNet concepts are frequently used as sources to provide factual and commonsense knowledge. There are two common approaches to utilize the knowledge. The first approach is parsing the knowledge in a symbolic format \cite{zhu2020mucko,narasimhan2018straight,narasimhan2018out,li2020boosting,marino2020krisp} that usually consists of a collection of (subject, relation, object) triplets. Although each triplet presents explicit knowledge, many conditions and context are lost when producing these triplets. The missing information could prevent VQA systems from learning whether the knowledge is valid for the question and disambiguating the entities in the triplet. On the contrary, the second approach relies on simply using free-form knowledge \cite{Wu_2016_CVPR,marino2019ok,dpr_okvqa} and use the raw text as input. While preserving most information, identifying helpful knowledge is quite challenging, especially in the multi-modal settings.  
\cite{Wu_2016_CVPR,marino2019ok} employ rule-based approaches that find knowledge relevant to one or a combination of objects detected in images or mentioned questions. \cite{dpr_okvqa} employ a dense retrieval approach that measures the relevance of the articles and the question-image pair in the feature space. However, being relevant to certain visual content cannot guarantee the helpfulness of the knowledge to predict the answer. To this end, we present answer-guided knowledge retrieval where the retrieved knowledge contains both the answer candidates and relevant visual content. Besides textual knowledge from Wikipedia and ConceptNet, we also explore external visual knowledge retrieved from the Google Images search engine.

\noindent \textbf{Answer Validation.}
The idea of using answer candidates has been used in various question answering settings, including textual QA \cite{zhang2021joint}, product QA \cite{zhang-etal-2020-answerfact}, commonsense VQA \cite{wu2020improving}, movie QA \cite{Kim2019ProgressiveAM}, allowing systems to perform a more in-depth examination of each answer candidate. We extend this idea to knowledge-based VQA, where the system accesses answer-specific external knowledge to assess the correctness of each answer candidate.

%%%%%%%%%%%%%%%%%%%%%%%%%

\begin{figure*}[!t]
    \centering
    \includegraphics[width=\linewidth,trim={0cm 29.8cm 42cm 0cm},clip]{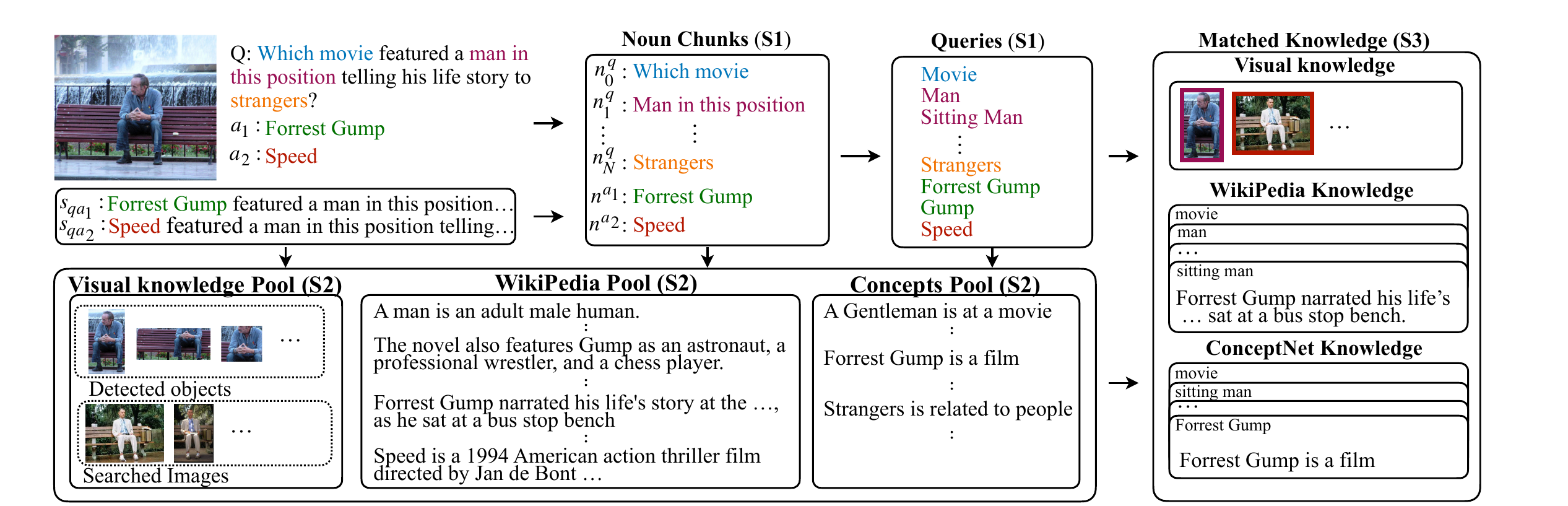}
    \caption{An example of the retrieval process for one question-answer pair. The numbers in parentheses denote the step number in Section \ref{sec:retrieval}. The noun phrase, its generated queries, and the matched visual knowledge are marked in the same color. }
    \label{fig:retrival}
\end{figure*}

\section{The \mavex Framework}
We present the \mavex framework, a three-stage scheme that first generates a set of promising answer candidates, retrieves knowledge guided by these answer candidates, and finally validates these answer candidates. Different from previous works \cite{garderes2020conceptbert,marino2020krisp} that utilize textual knowledge, we propose to mine \emph{multi-modal answer-specific} knowledge. In particular, we consider three knowledge sources: Wikipedia and ConceptNet for text, and Google for images. These provide factual, commonsense, and visual knowledge, respectively. For validation, we test each answer candidate using the retrieved multi-modal knowledge.

\subsection{Answer Candidate Generation}
In order to use answer candidates to inform knowledge retrieval, we use ViLBERT-multi-task system \cite{lu2019vilbert}, a state-of-the-art VQA model, to generate answer candidates. In particular, we finetune a ViLBERT-multi-task model on the OK-VQA dataset that outputs a score for each answer collected from the training set. The highest-scoring answers are used as the candidates. Note that any VQA model or other approaches (for example, querying ontology knowledge bases) can be used for this purpose. However, as we will discuss in the experiments section, we found ViLBERT to be particularly effective at generating a small set of promising candidates. 
%%%%%%%%%

\subsection{Answer Guided Knowledge Retrieval}
\label{sec:retrieval}
Given a question $q$ about an image $I$ and a set of answer candidates $A$, we retrieve external knowledge supporting $A$ in three main steps.
Figure~\ref{fig:retrival} shows the entire process for an example question and a candidate answer.

\subsubsection{S1: Query Extraction.} 
We first collect short phrases in $q$, each answer candidate in $A$, and concepts represented in $I$ as a starting point for retrieving external information. This involves the following sub-steps:

\underline{Extract noun phrases from question and answers:}
We parse the question and the candidate answers using a constituency parser to obtain the parse tree. Then, we extract all the nouns on the leaves of the parse tree together with the words that describe the nouns and belong to one of the types from $\texttt{ADJP}$, $\texttt{ADVP}$, $\texttt{PP}$, $\texttt{SBAR}$, $\texttt{DT}$ or $\texttt{JJ}$. 
We extract three kinds of noun phrases for modeling: 
(1) The target noun phrase that contains `wh' or `how' word (e.g. \emph{`which movie'}), denoted by $n^q_0$. (2) Question noun phrases from the rest of the question (e.g., \emph{`man in this position'}), denoted by $n^q_i, i \in \{1,\ldots,N\}$. $N$ is the number of noun phrases in the question. (3) Answer noun phrase for each answer $a_i$, denoted by $n^{a_i}$ (e.g., \emph{`Forrest Gump'}). These nouns help us link the mentioned objects to the images. 

\underline{Link phrases to objects:}
As images usually contain plenty of question-irrelevant content, making the retrieval process hard, we propose narrowing the search to the objects referred to by the question or the answer candidates. In particular, we use  a separate ViLBERT-multi-task \cite{Lu_2020_CVPR} model as the object linker, where it takes as inputs a set of detected objects and a noun phrase from the question, and outputs a linking score for each detected object to indicate how likely the noun phrase refers to the object.
We approve the linking when the score is higher than 0.5 and extract the linked objects. 

\underline{Generate search query set:} 
We further generate a set of search queries to search the external knowledge base. For each noun phrase, we first extract the head of a phrase by finding the innermost \texttt{NP} from the dependency tree. 
%(e.g., ``man'' in the phrase ``man in this position'') 
Then, we obtain the visual attributes of the head of the noun phrase by using a pre-trained object-with-attribute detector \cite{anderson2017bottom} for the corresponding linked objects. For example, the visually grounded queries for \emph{`man in this position'} are \emph{`man'} and \emph{`sitting man'} where \emph{sitting} is inferred from visual attributes. 
We denote the set of queries as $r^n_i, i \in \{1,\ldots,K\}$, where $n$ is the corresponding noun phrase, $K$ is the maximum number of queries per noun phrase.

\subsubsection{S2: Answer Guided Knowledge Pool Construction.} We now use the visually grounded queries from step S1 to construct knowledge pool as follows:

\underline{Conversion to a natural language statement:}
In order to use the answer candidate $a$ to inform the retrieval step, we convert $q$ and $a\in A$ into a natural language statement $s_{qa}$ using a T5 model \cite{2020t5} finetuned on the QA-NLI dataset \cite{demszky2018transforming}. Such conversion is effective as statements occur much more frequently than questions in textual knowledge sources~\cite{Khot2017AnsweringCQ}. These statements are later used to compute the relevance of the retrieved facts as described below. 

\underline{Retrieval of textual facts and concepts:} We search each query in the query set generated from the last sub-step in \textbf{S1} in Wikipedia and ConceptNet. We compute the BERTScore \cite{zhang2019bertscore} between each sentence from the retrieved article and each statement $s_{qa}$. For each statement, the top-$15$ sentences (according to the BERTScore) from each retrieved article are pushed to the sentence pool. Then, we decontextualize \cite{choi2021decontextualization} each sentence in the Wikipedia pool for better knowledge quality.

\underline{Retrieval of visual knowledge:} Pure textual knowledge is often insufficient due to two main reasons: (1) textual knowledge might be too generic and not specific to the question image, (2) it might be hard to describe some concepts using text, and an image might be more informative (e.g., the third question in Figure~\ref{fig:teaser}).  Hence, visual knowledge can complement textual information, further enriching the external knowledge space. 
We consider both internal and external visual knowledge. For the given image, we utilize a MaskRCNN \cite{he2017mask} object detector to detect common objects as internal knowledge. We use Google image search to retrieve the top-5 images using the statement $s_{qa}$ as the query for each answer candidate $a$ as the external visual knowledge. 

\subsubsection{S3: Matching Knowledge Pool to Queries.} 
Instead of simply using each query's top retrieved sentences as the query's knowledge, we propose matching the sentences from the entire pool to each query. The intuition is that most queries cannot directly retrieve helpful facts; however, they can help retrieve important aspects that should be contained in the external knowledge. 

\underline{Matching Textual Knowledge:} 
For each query, the sentences from both Wikipedia and ConceptNet pool with a mean recall greater than 0.6 are considered the retrieved results. Mean recall is the average cosine similarity between the Glove embedding of the words in the query and their most similar word in the sentence. To ensure knowledge relevance, we remove sentences that are matched to only a single query. 
For each query $r_i^n$, according to the maximum BERTScore between the sentence and all of the statements $S_{q}$, we extract at most $m$ sentences from both Wikipedia and ConceptNet pools, denoted by $W(r_i^n)$ and $C(r_i^n)$.

\underline{Matching Visual Knowledge:} 
For each noun phrase in the question, we can directly use the results from the object linker defined in S1. Specifically, we find the top-$3$ referenced objects in the image for each question noun phrase, denoted $M(n)$.

For each answer noun phrase $n^{a_i}$, we use Google image search to retrieve the top-$5$ images, denoted $M(n^{a_i})$.

\begin{figure*}[!t]
    \centering
    \includegraphics[width=38pc,trim={0cm 28cm 43cm 0cm},clip]{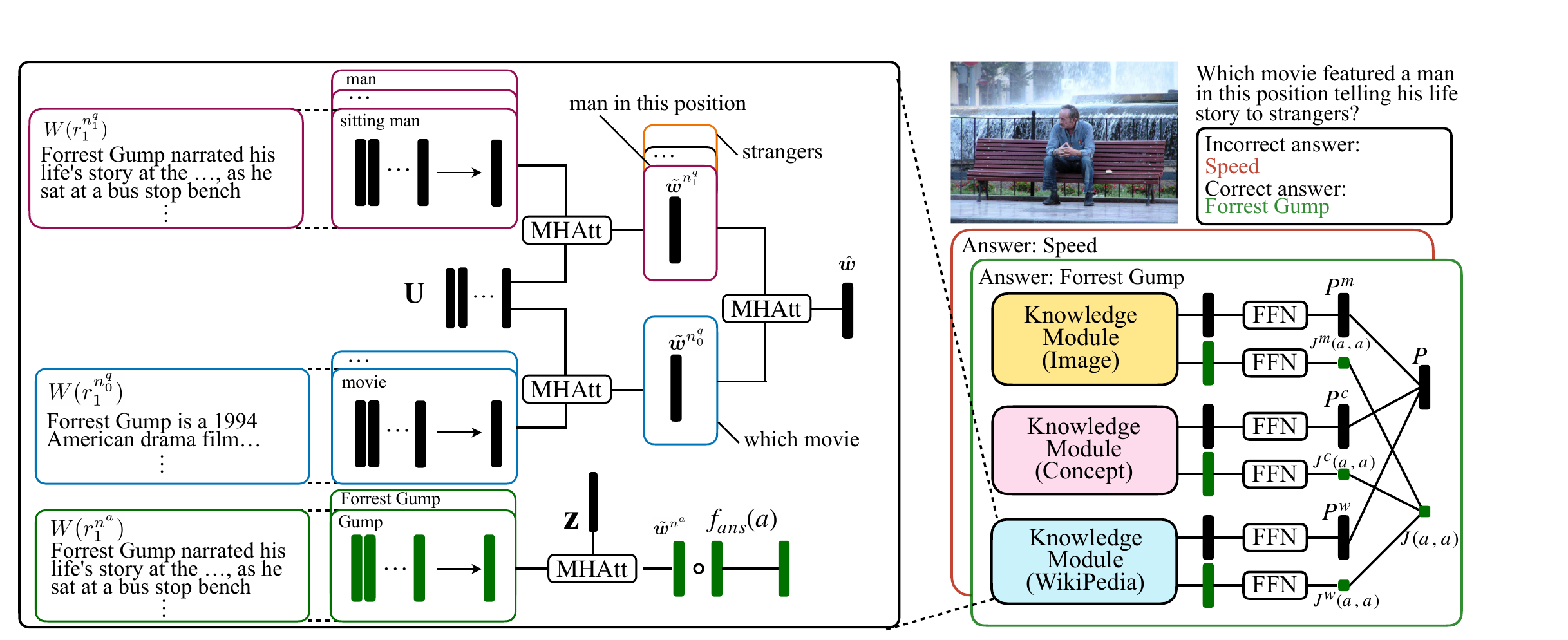}
    \caption{Model overview for validating two candidate answers. We explore three sources of external knowledge, $i.e.$ Wikipedia, ConceptNet, and Google Images presented by the three parallel knowledge embedding modules. The black blocks denote features shared by all answer candidates, and the green blocks denote answer-specific features. 
    Different colors denotes the features for different noun phrases and their queries. }
    \label{fig:model}
\end{figure*}

\subsection{Answer Candidate Validation}
\label{sec:valid}
The answer validation module takes as input an answer candidate $a_i$ and the supporting knowledge, and outputs a scalar score indicating how well the knowledge supports $a_i$.  As we will discuss, in order to better aggregate the knowledge, we first compute the knowledge embedding for each query. Then, we compute an embedding for each noun phrase that aggregates the embedding for the queries generated from the noun phrase\footnote{Recall that our queries $r_i^n$ are created based on noun phrase $n$.}. Finally, the embedding for the entire question aggregates the embedding computed for all noun phrases.

% talk about feature extraction here.
We build \mavex on top of the ViLBERT system. Given a question $q$ and an image $I$, ViLBERT provides textual features $\bm{U}\in \mathbb{R}^{|q| \times d}$, visual features $\bm{V}\in \mathbb{R}^{|V| \times d}$ from the last layer, where $|q|$ is the number of tokens in $q$, $d$ is the feature dimension, $|V|$ is the number of objects in the image plus one for the representation for the entire image, and a joint visual-textual representation $\bm{z}\in \mathbb{R}^{d}$. 
For each sentence in the retrieved textual knowledge $W(r_i^n)$ and $C(r_i^n)$, we use TinyBERT (T-BERT) model \cite{turc2019well} to extract the corresponding features. We further average the sentence features for each query $r_i^n$, resulting $\bm{w}^n_i$ and $\bm{c}^n_i$.

For each image in the retrieved visual knowledge $M(n_a)$, we use MaskRCNN \cite{he2017mask} to extract object features. Then, we average the object features of visual detection results as the image features and denote them as $\bm{m}^n_i$. Note that we directly use the object features for the linked objects. 
Figure \ref{fig:model} shows the overview of the model.

\subsubsection{Multi-Granular Knowledge Embedding Module.}

In order to better aggregate the retrieved knowledge, we employ a multi-granular knowledge embedding module that learns to recognize the critical queries for each noun phrase, and then the critical noun phrases for answering the question.

Note that our knowledge embedding module is identical for each knowledge source but with different learnable parameters. We only show knowledge from Wikipedia for brevity. 
Given the knowledge embeddings $\bm{w}^n_i$ for each query $r^n_i$ in the question, we compute the knowledge embedding $\tilde{\bm{w}}^n$ for each noun phrase in question as follows:
\begin{equation}
    \tilde{\bm{w}}^n = \texttt{MHAtt}(\bm{u}^n, \{\bm{w}^n_i\}_{i \in \{1,\ldots,K\}},  \{\bm{w}^n_i\}_{i \in \{1,\ldots,K\}} ),
    % \texttt{Adding an equation here.}
\end{equation}
where $\texttt{MHAtt}(\texttt{query},\texttt{key},\texttt{value})$ is the multi-head attention operator. $\bm{u}^n$ is the attentive pooled \cite{lee2017end} ViLBERT features according to the span $\{s, e\}$ of the phrase $n$. 
% explain more here.
We use $\bm{u}^n$ as the \texttt{query} in \texttt{MHAtt} module to aggregate the retrieved knowledge, where the corresponding knowledge embeddings $\{\bm{w}^n_i\}_{i \in \{1,\ldots,K\}}$ serve as \texttt{key} and \texttt{value}. 

Similarly, for each answer $a$\footnote{For simplicity we omit the subscript index of the answer in this section when there is only one answer involved in the current step.}, we compute the knowledge embedding $\bm{w}^{a}$ using a $\texttt{MHAtt}$ module over the knowledge features $\bm{w}^{n_a}_i$ as follows:
\begin{equation}
    \bm{w}^{a} = \texttt{MHAtt}(\bm{z}, \{\bm{w}^{n_a}_i\}_{i \in \{1,\ldots,K\}},  \{\bm{w}^{n_a}_i\}_{i \in \{1,\ldots,K\}} ),
    % \texttt{Adding an equation here.}
\end{equation}
where the joint visual-textual embeddings $\bm{z}$ from ViLBERT system is used as the keys.

Then, another $\texttt{MHAtt}$ module is used to gather the knowledge from each noun phrase $n \in \{n^q_1, \ldots,n^q_N\}$. Specifically, given the knowledge embedding for each noun phrase, the knowledge embeddings $\bm{w}$ is computed as follows: 
%\todo{I don't understand why there are two $n$ indices}
\begin{equation}
    \hat{\bm{w}} = \texttt{MHAtt}(\tilde{\bm{w}}^{n^q_0}, \{\tilde{\bm{w}}^{n^q_i}\}_{i \in 1,\ldots,N}, \{\tilde{\bm{w}}^{n^q_i}\}_{i \in 1,\ldots,N})
    % \texttt{Adding an equation here.}
\end{equation}

\subsubsection{Answer Prediction and Validation Module}
Given the knowledge embedding $\bm{k} \in \{\hat{\bm{w}}, \  \hat{\bm{c}}, \  \hat{\bm{m}}\}$ from each one of the three knowledge sources, \mavex predicts the answers' probability as $P^{\bm{k}}=\texttt{FFN}(\bm{k} + \bm{z})$, where $\texttt{FFN}$ denotes a feed-forward layer. The final prediction $P$ is the answer that has the maximum confidence over the three knowledge sources for each answer, $i.e.$ $P = \max\limits_{\bm{k}} \{ P^{\bm{k}} \}$.

The validation module takes as inputs the answer candidate $a$ and the knowledge features $\bm{k}^{\primemacro{a}} \in \{\bm{w}^{\primemacro{a}}, \bm{c}^{\primemacro{a}}, \bm{m}^{\primemacro{a}}\}$ from the three sources to learn how well the knowledge supports the answer candidate. We first embed the answer candidate using the summation of the BERT features of the corresponding statement and the glove features of the answer itself, $i.e.$ $f_{ans}(a) = (\text{BERT}(s_{qa}) + \text{glove}(a))$. Then, the validation score $J(a, \primemacro{a})$ for answer candidate $a$ using the knowledge retrieved for $\primemacro{a}$ (a different candidate) is computed as $J^{\bm{k}}(a, \primemacro{a}) = \text{FFN}(f_{ans}(a) \circ \bm{k}^{\primemacro{a}})$, where the $\circ$ means element-wise multiplication. The final validation score is the maximum validation confidence over the three knowledge sources, $i.e.$ $J(a, \primemacro{a}) =\max\limits_{\bm{k}} \{J^{\bm{k}}(a, \primemacro{a})\} $.

\begin{table*}[t]
\centering
\begin{tabular}{l|c|c}
\toprule
\textbf{Method}  & \textbf{Knowledge Resources} & \textbf{Performance} \\\hline
ArticleNet (AN) \cite{marino2019ok}  & Wikipedia &  \phantom{0}5.3 \\
Q-only \cite{marino2019ok}  & --- &  14.9   \\
MLP \cite{marino2019ok}  & --- &   20.7 \\ \cdashline{3-3}
BAN \cite{kim2018bilinear}  & --- & 25.2   \\
\ \ \ \ + AN \cite{marino2019ok} & Wikipedia &  25.6   \\
\ \ \ \ + KG-AUG \cite{li2020boosting} & Wikipedia + ConceptNet & 26.7 \\ \cdashline{3-3}
MUTAN \cite{ben2017mutan}  & --- &  26.4 \\
\ \ \ \ + AN \cite{marino2019ok} & Wikipedia & 27.8  \\ \cdashline{3-3}
Mucko \cite{zhu2020mucko}  &  Dense Caption  &  29.2  \\ 
ConceptBert \cite{garderes2020conceptbert}  &  ConceptNet  &  33.7 \\
KRISP \cite{marino2020krisp}  & Wikipedia +  ConceptNet  &  38.9$^{*}$ \\  
RVL$^\dag$ \cite{shevchenko2021reasoning}  &  Wikipedia + ConceptNet  &  39.0$^\dag$ \\
\mavex (ours)  & Wikipedia + ConceptNet & 39.45$^{*}$ \\
\mavex (ours)  & Wikipedia + ConceptNet + Google Images  & \textbf{40.28}$^{*}$ \\ \hline
\mavex (ours)  (Ensemble 3) & Wikipedia + ConceptNet + Google Images  & \textbf{41.37}$^{*}$ \\
\bottomrule
\end{tabular}
\caption{\mavex outperforms current state-of-the-art approaches on OK-VQA. The middle column lists the external knowledge sources, if any, used in each system. \dag\ indicates that the system uses a pretrained model contaminated by OK-VQA test images. $^*$ indicates that the results have been reported on version 1.1 of the dataset.}
\label{tab:vqa_results}
\end{table*}

\noindent \textbf{Consistency Criteria.}
The intuition behind our consistency criteria is that for the correct answer $a$, the knowledge retrieved for $a$ from the most confident source (the one with the highest supportiveness score $J$ for $a$) should support $a$ more than it supports other answer candidates, and it should also support $a$ more than knowledge retrieved for other answer candidates. Specifically, we approve the answer validation score $J(a, a)$ only if it is higher than the scores computed using this knowledge for all other answers as well as the score for $a$ when using knowledge retrieved for other answers. We also eliminate the case where the top-1 prediction from $P$ is not in the answer candidate set. Mathematically, the consistency criteria checks that $J(a, a) > J(\primemacro{a}, a)$ and $J(a, a) > J(a, \primemacro{a})$ for all $\primemacro{a} \neq a$. If the above condition is not met, we output the answer with the maximum VQA prediction score $P(a)$; otherwise, we output the answer with the maximum VQA-weighted validation score $J(a, a) P(a)$. 

\subsection{Training and Implementation Details}
% We present the training and implementation details in this section.\\

\noindent\textbf{Implementation.} We implemented our approach on top of  ViLBERT-multi-task \cite{lu2019vilbert}, which utilizes a Mask-RCNN head \cite{he2017mask} in conjunction with a ResNet-152 base network \cite{he2016deep} as the object detection module. Convolutional features for at most $100$ objects are then extracted for each image as the visual features, $i.e.$ a 2,048-dimensional vector for each object. We used the constituent parser from AllenNLP to extract the nouns phrases in the question. For linking the mentioned objects, we adopt a separate ViLBERT-multi-task system. For converting the question and answer, we finetuned a T5-base model \cite{2020t5} on the QA-NLI dataset \cite{demszky2018transforming} for 4 epochs. We detected 100 objects using Mask-RCNN to encode the retrieved Google images. For question embedding, following \cite{devlin2018bert}, we use a BERT tokenizer on the question and use the first $23$ tokens as the question tokens. We encode at most $4$ sentences per query, $3$ queries per noun phrase. The number of hidden units in the multi-head attention modules is set to $512$. We use Pytorch 1.4 on a single TITAN V GPU with 12M memory for each run, and it generally costs 22 hours to train a single model.\\

\noindent\textbf{Training.}
%\todo{moved training after implementation}
The OK-VQA test images are a subset of COCO validation images that are used to pre-train most transformer-based vision and language models  \cite{lu2019vilbert,tan2019lxmert,li2019visualbert}. Although the test questions never appear in the pre-training process, other questions on the test images may help the system understand the image better, leading to higher performance. Besides, there is also data contamination from extra object annotations from Visual Genome (VG) dataset, which also contains some OK-VQA test images. As the VG dataset is used to pre-train the object detector, those test images can access the ground truth object annotations. Therefore, we carefully remove all OK-VQA test images from the pre-training and re-train the ViLBERT-multi-task model and the object detector from scratch using the default configurations.

We finetune the ViLBERT-multi-task model on OK-VQA using the default configuration for $150$ epochs for answer candidate generation. Binary cross-entropy loss and VQA soft score are employed to optimize the system. OK-VQA provides five annotations for each question. Soft scores are $0$, $0.6$, and $1$ corresponding to $0$, $1$, more than $1$ matching answer annotations. We use the finetuned model to extract the top $5$ answers for each question in the training and test set. We follow the default settings of ViLBERT and apply the BertAdam optimizer \cite{devlin2018bert} with a linear warmup learning rate. 

For the training of the answer validation module, we optimize the validation score $J(a, \primemacro{a})$ using the loss in Eq. \ref{eq:loss_mavex} for the three knowledge sources, where $s(a)$ %\todo{I think it was P(a) in the previous section (P(a) is the output and S(a) is the supervision} 
denotes the VQA soft scores for answer $a$. We also add the standard VQA losses on the predictions from the three external sources. We train the system for 75 epochs using a learning rate of $2$e-$5$ for the ViLBERT parameters and $5$e-$5$ for the additional parameters introduced in the validation module. We freeze the first $10$ layers of the ViLBERT base network. We use $\mathcal{L}_{bce}$ to denote binary cross-entropy loss.

\begin{align}
\nonumber   \mathcal{L}_{\text{\mavex}} & =  \mathcal{L}_{bce} \Big(\max_{\begin{subarray}{c}\text{$a$}\\\text{$s.t.  \ a \neq \primemacro{a}$ }\end{subarray}} J(a, \primemacro{a}), \ \textbf{0} \Big)  \\
\nonumber    &+  \mathcal{L}_{bce}\Big(   \max_{\begin{subarray}{c}\text{$\primemacro{a}$}\\ \text{$s.t.  \ a \neq \primemacro{a}$ }\end{subarray}} J(a, \primemacro{a}), \ \textbf{0}\Big) \\
 &+ \mathcal{L}_{bce}\Big(J(a, a), s(a)\Big) \label{eq:loss_mavex}
\end{align}

\section{Experiments}
We evaluate our framework on the OK-VQA dataset. We first briefly describe the dataset and then present our results, comparing with the current state-of-the-art systems. \\

\noindent\textbf{OK-VQA dataset}~\cite{marino2019ok} is the largest knowledge-based VQA dataset at present. The questions are crowdsourced from Amazon Mechanical Turkers, leading to two main advantages: (1) the questions indeed require outside knowledge beyond images; (2) there are no existing knowledge bases that cover all the questions, thus requiring systems to explore open-domain resources. The dataset contains 14,031 images and 14,055 questions covering a variety of knowledge categories. The metric is the VQA soft score. For the experiments of this paper, we used version 1.1 of the dataset.

\subsection{Intrinsic Evaluation}
We begin with an intrinsic evaluation of \mavex, assessing the quality of the answer candidate generation step.\\ %and knowledge retrieval modules.\\

\noindent\textbf{Answer Candidate Accuracy.}
Our answer candidate generation module, based on the finetuned ViLBERT-multi-task model, outputs its top-$5$ answers as the candidates. We found that the best answer in this small candidate set achieves a VQA soft score of $59.7$ on the test set, substantially higher than other state-of-the-art systems without data contamination. We also evaluate the score achieved by slightly larger candidate sets, consisting of the top 6, 8, and 10 candidates. These achieve VQA soft scores of $62.1$, $65.1$, and $67.1$, respectively. Since our answer validation framework needs to retrieve and encode answer-specific knowledge, we use only top-$5$ answer candidates as a reasonable trade-off between efficiency, answer coverage, and overall accuracy. Note that our method cannot produce answers not in the candidate set.

\begin{figure}[ht]
    \centering
    \includegraphics[width=19pc]{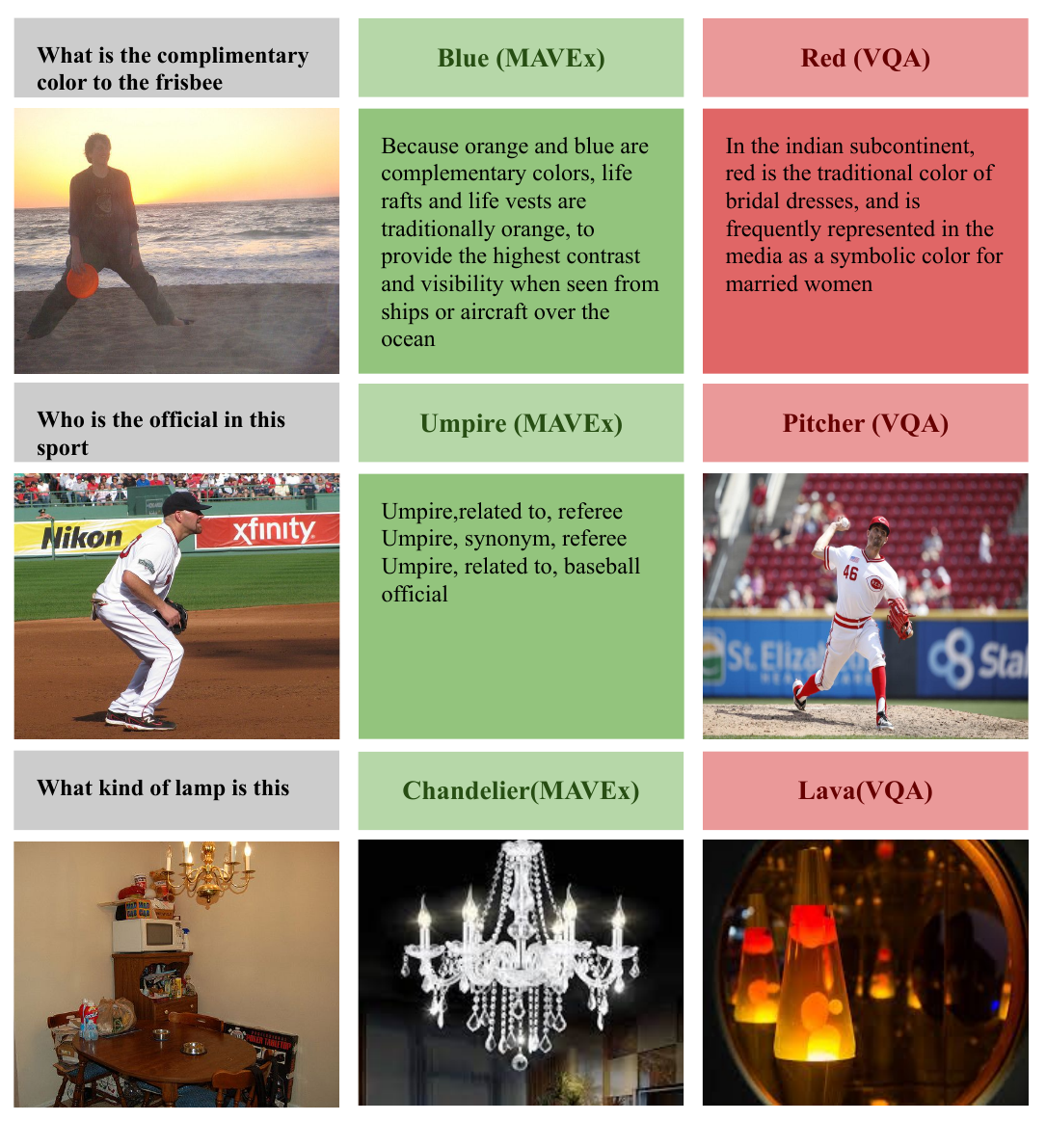}
    \caption{Examples where the VQA model is wrong but \mavex with the three external knowledge sources answers correctly. The correct answer is in the green box and the incorrect answer is shown in the red box. The grey box shows the question. Sample retrieved knowledge content is shown in the boxes under the predicted answers. }
    \label{fig:qe1}
\end{figure}

\subsection{Main Results}
Table~\ref{tab:vqa_results} shows that \mavex consistently outperforms prior approaches by a clear margin. For example, \mavex single model outperforms recent state-of-the-art models, KRISP~\cite{marino2020krisp}, and ConceptBert~\cite{garderes2020conceptbert} by 1.4, 6.6 points, respectively. An ensemble of three \mavex  models with different initializations provides 2.47 points improvement compared to KRISP. All of our results, except for the ensemble model, are averaged across $3$ different initialization seeds. The standard deviation is 0.21 for the single model computed from the three runs.

\subsection{Ablation Studies}

\noindent \textbf{Knowledge Sources.}
We report the performance of using the different combinations of knowledge sources in Table~\ref{tab:oracle}. We see that the three sources (WikiPedia, ConceptNet, and Images) improve the performance by $3.4$, $3.3$, and $3.1$, respectively, compared to the base ViLBERT system. This indicates the effectiveness and value of all three sources. The decontextualization technique \cite{choi2021decontextualization} improves the performance compared to using only the Wikipedia source by 0.4\%. The decontextualization partially helps address the co-reference issue since the retrieved sentences provide more information from their paragraph. Combining the three sources achieves a net performance gain of $5\%$ over the ViLBERT baseline, supporting the intuition that the three sources together provide complementary pieces of knowledge. 

We show some qualitative examples in Figure~\ref{fig:qe1}, where the VQA model (ViLBERT) is wrong but provides good answer candidates. Our \mavex gathers the external knowledge from the three sources and predicts the correct answers.

\begin{table}[ht]
\centering
\begin{tabular}{l|c}
\toprule
 & System \\
Knowledge Source & Score  \\ \hline
 ViLBERT & $35.20$    \\
 Wikipedia (\textit{w/o} decontextualization) & $38.21$    \\
 Wikipedia & $38.63$    \\
 ConceptNet & $38.56$         \\
  Images & $38.30$      \\
 Wikipedia + ConceptNet & $39.45$  \\
 ConceptNet + Images& $39.60$       \\
  Wikipedia +  Images & $39.37$     \\
 Wikipedia + ConceptNet + Images  & $40.28$\\ 
 Wikipedia + ConceptNet + Images (Oracle) & $47.76$\\ 
 \bottomrule
\end{tabular}
\caption{Ablation study using different combinations of knowledge sources.}
\label{tab:oracle}
\end{table}

% \subsection{Ablation of knowledge embedding granularity}
\noindent \textbf{Knowledge Embedding Granularity.}
We ablate different levels of granularity used in the \mavex system by comparing to two baseline systems, where we replace the noun-phrase level or the question level 
%\todo{not sure what you mean by first level and second level (changed to noun-phrase level or the question level)} 
multi-head attention module with an average pooling operation. When replacing the noun-phrase level $\texttt{MHAtt}$ modules ($i.e.$ three $\texttt{MHAtt}$ modules corresponding to merging queries' features for question noun phrases, question target phrase and the answer phrases), the  performance reduces to $39.77$. When replacing the question level $\texttt{MHAtt}$ module with an average pooling operation for the question, the performance reduces to $39.60$.

\noindent \textbf{Answer Validation Step.}
We consider a \mavex baseline model that uses the retrieved knowledge ($\tilde{\bm{w}}$, $\tilde{\bm{c}}$, $\tilde{\bm{m}}$) as additional inputs but without answer validation. This model achieves an overall score of 39.2, 4\% higher than the ViLBERT base model and 1.1\% lower than the full model, indicating that using answer-guided retrieved knowledge is helpful, and answer validation further improves performance.

\subsection{Oracle Experiments}

\noindent\textbf{Oracle Source Selector.} 
We report an oracle score obtained by manually choosing the best verification score $J^{\bm{k}}(a, a)$ from the three sources $\bm{k} \in \{\tilde{\bm{w}}, \ \tilde{\bm{c}}, \ \tilde{\bm{m}}\}$ to weigh the prediction $P$. As a result, our answer validation framework achieves an oracle score of 47.76 as reported in Table \ref{tab:oracle}. This indicates that the three knowledge sources provide complementary features, leaving further potential to improve the system.

\begin{figure}[t]
    \centering
    \includegraphics[width=\columnwidth]{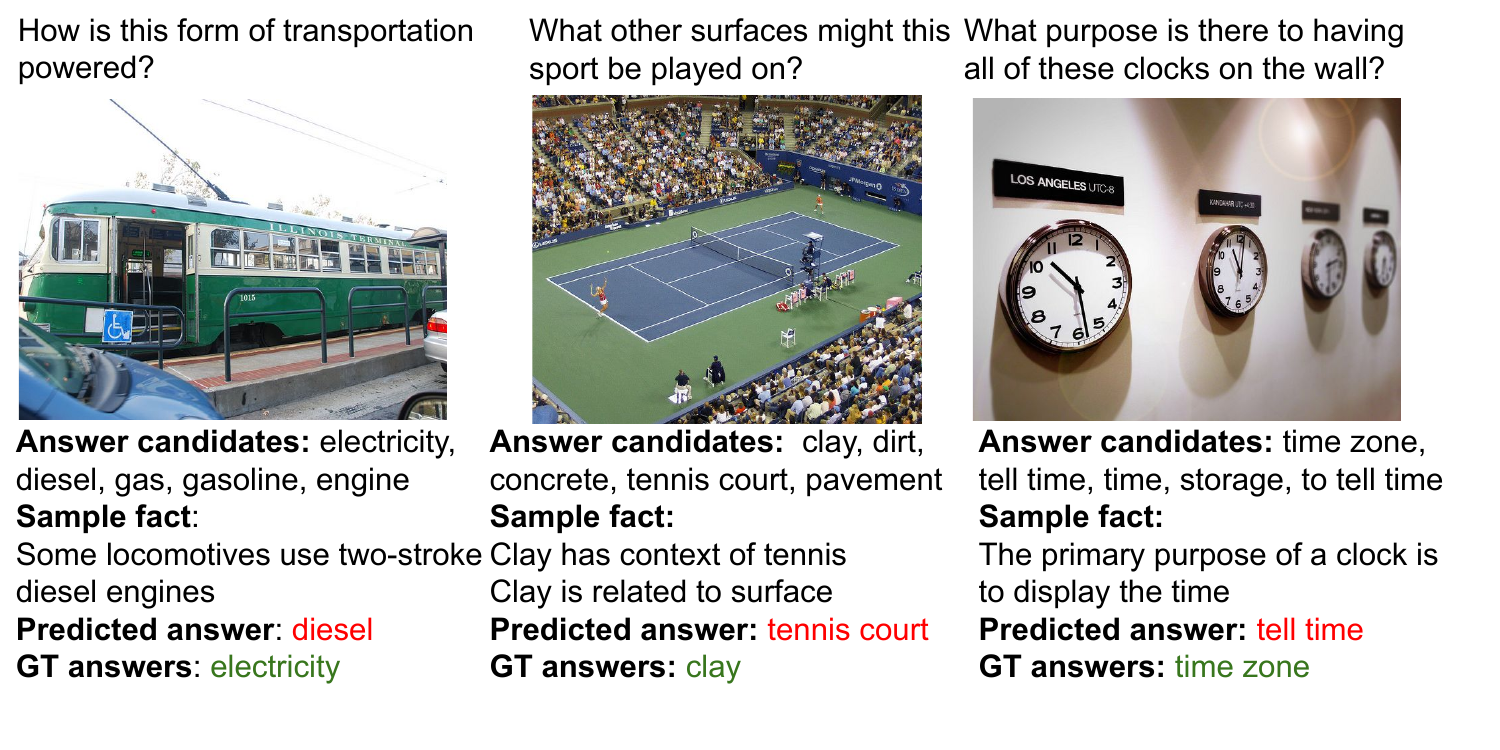}
    \caption{Some typical failure cases of our model have been shown. In these examples, the model falsely focuses on the retrieved fact (left), visual content (middle), or does not generate proper search word (right).}
    %\vspace{-0.3cm}
    \label{fig:rebuttal}
\end{figure}

\subsection{Failures Cases Analysis} 
Figure~\ref{fig:rebuttal} shows some common types of failure examples. In the left example, the model over-relies on the retrieved fact ``some locomotives use diesel engines'' and ignores the key visual clue in the image (the wires above the train). In the middle example, the model relies on the visual content ``tennis court'' and does not use the retrieved knowledge. In the example shown on the right, the model fails to realize that the key clue is the \textit{difference in displayed time on the clocks}.

\section{Conclusion}
We presented \mavex, a novel approach for knowledge based visual question answering. The goal is to retrieve answer-specific textual and visual knowledge from different knowledge sources and learn what sources contain the most relevant information. Searching through the vast amount of retrieved knowledge, which is often quite noisy, is challenging. Hence, we formulate the problem as answer validation, where the goal is to learn to verify the validity of a set of candidate answers according to the retrieved knowledge. More specifically, an answer candidate validation module predicts the degree of support provided by the knowledge retrieved for each candidate, and decides which sources to trust for each candidate answer. \mavex demonstrates the clear advantages of answer-guided knowledge retrieval, achieving the state-of-the-art performance on the OK-VQA dataset.

\bibliography{egbib}

\begin{thebibliography}{46}
\providecommand{\natexlab}[1]{#1}

\bibitem[{Anderson et~al.(2018)Anderson, He, Buehler, Teney, Johnson, Gould,
  and Zhang}]{anderson2017bottom}
Anderson, P.; He, X.; Buehler, C.; Teney, D.; Johnson, M.; Gould, S.; and
  Zhang, L. 2018.
\newblock Bottom-{U}p and {T}op-{D}own {A}ttention for {I}mage {C}aptioning and
  {VQA}.
\newblock In \emph{CVPR}.

\bibitem[{Antol et~al.(2015)Antol, Agrawal, Lu, Mitchell, Batra,
  Lawrence~Zitnick, and Parikh}]{antol2015vqa}
Antol, S.; Agrawal, A.; Lu, J.; Mitchell, M.; Batra, D.; Lawrence~Zitnick, C.;
  and Parikh, D. 2015.
\newblock {VQA: Visual Question Answering}.
\newblock In \emph{ICCV}.

\bibitem[{Ben-Younes et~al.(2017)Ben-Younes, Cadene, Cord, and
  Thome}]{ben2017mutan}
Ben-Younes, H.; Cadene, R.; Cord, M.; and Thome, N. 2017.
\newblock {MUTAN: Multimodal Tucker Fusion for Visual Question Answering}.
\newblock In \emph{ICCV}.

\bibitem[{Cadene et~al.(2019)Cadene, Ben-Younes, Cord, and
  Thome}]{cadene2019murel}
Cadene, R.; Ben-Younes, H.; Cord, M.; and Thome, N. 2019.
\newblock Murel: Multimodal relational reasoning for visual question answering.
\newblock In \emph{CVPR}.

\bibitem[{Chen et~al.(2020)Chen, Li, Yu, Kholy, Ahmed, Gan, Cheng, and
  Liu}]{chen2019uniter}
Chen, Y.-C.; Li, L.; Yu, L.; Kholy, A.~E.; Ahmed, F.; Gan, Z.; Cheng, Y.; and
  Liu, J. 2020.
\newblock Uniter: Universal image-text representation learning.
\newblock In \emph{ECCV}.

\bibitem[{Choi et~al.(2021)Choi, Palomaki, Lamm, Kwiatkowski, Das, and
  Collins}]{choi2021decontextualization}
Choi, E.; Palomaki, J.; Lamm, M.; Kwiatkowski, T.; Das, D.; and Collins, M.
  2021.
\newblock Decontextualization: Making Sentences Stand-Alone.
\newblock \emph{TACL}.

\bibitem[{Demszky, Guu, and Liang(2018)}]{demszky2018transforming}
Demszky, D.; Guu, K.; and Liang, P. 2018.
\newblock Transforming question answering datasets into natural language
  inference datasets.
\newblock \emph{arXiv}.

\bibitem[{Devlin et~al.(2019)Devlin, Chang, Lee, and
  Toutanova}]{devlin2018bert}
Devlin, J.; Chang, M.-W.; Lee, K.; and Toutanova, K. 2019.
\newblock Bert: Pre-training of deep bidirectional transformers for language
  understanding.
\newblock In \emph{NAACL-HLT}.

\bibitem[{Gard{\`e}res et~al.(2020)Gard{\`e}res, Ziaeefard, Abeloos, and
  Lecue}]{garderes2020conceptbert}
Gard{\`e}res, F.; Ziaeefard, M.; Abeloos, B.; and Lecue, F. 2020.
\newblock ConceptBert: Concept-Aware Representation for Visual Question
  Answering.
\newblock In \emph{EMNLP}.

\bibitem[{He et~al.(2017)He, Gkioxari, Doll{\'a}r, and Girshick}]{he2017mask}
He, K.; Gkioxari, G.; Doll{\'a}r, P.; and Girshick, R. 2017.
\newblock Mask r-cnn.
\newblock In \emph{ICCV}.

\bibitem[{He et~al.(2016)He, Zhang, Ren, and Sun}]{he2016deep}
He, K.; Zhang, X.; Ren, S.; and Sun, J. 2016.
\newblock {Deep Residual Learning for Image Recognition}.
\newblock In \emph{CVPR}.

\bibitem[{Hudson and Manning(2019)}]{hudson2019gqa}
Hudson, D.~A.; and Manning, C.~D. 2019.
\newblock GQA: a new dataset for compositional question answering over
  real-world images.
\newblock In \emph{CVPR}.

\bibitem[{Kazemzadeh et~al.(2014)Kazemzadeh, Ordonez, Matten, and
  Berg}]{refcoco}
Kazemzadeh, S.; Ordonez, V.; Matten, M.; and Berg, T. 2014.
\newblock {R}efer{I}t{G}ame: Referring to Objects in Photographs of Natural
  Scenes.
\newblock In \emph{EMNLP}.

\bibitem[{Khot, Sabharwal, and Clark(2017)}]{Khot2017AnsweringCQ}
Khot, T.; Sabharwal, A.; and Clark, P. 2017.
\newblock Answering Complex Questions Using Open Information Extraction.
\newblock In \emph{ACL}.

\bibitem[{Kim et~al.(2019)Kim, Ma, Kim, Kim, and Yoo}]{Kim2019ProgressiveAM}
Kim, J.; Ma, M.; Kim, K.; Kim, S.; and Yoo, C.~D. 2019.
\newblock Progressive Attention Memory Network for Movie Story Question
  Answering.
\newblock In \emph{CVPR}.

\bibitem[{Kim, Jun, and Zhang(2018)}]{kim2018bilinear}
Kim, J.-H.; Jun, J.; and Zhang, B.-T. 2018.
\newblock {Bilinear Attention Networks}.
\newblock In \emph{NeurIPS}.

\bibitem[{Lee et~al.(2017)Lee, He, Lewis, and Zettlemoyer}]{lee2017end}
Lee, K.; He, L.; Lewis, M.; and Zettlemoyer, L. 2017.
\newblock End-to-end neural coreference resolution.
\newblock \emph{arXiv}.

\bibitem[{Li et~al.(2020)Li, Duan, Fang, Gong, Jiang, and
  Zhou}]{li2020unicoder}
Li, G.; Duan, N.; Fang, Y.; Gong, M.; Jiang, D.; and Zhou, M. 2020.
\newblock Unicoder-VL: A Universal Encoder for Vision and Language by
  Cross-Modal Pre-Training.
\newblock In \emph{AAAI}.

\bibitem[{Li, Wang, and Zhu(2020)}]{li2020boosting}
Li, G.; Wang, X.; and Zhu, W. 2020.
\newblock Boosting Visual Question Answering with Context-aware Knowledge
  Aggregation.
\newblock In \emph{ACM Conference on Multimedia}.

\bibitem[{Li et~al.(2019)Li, Yatskar, Yin, Hsieh, and Chang}]{li2019visualbert}
Li, L.~H.; Yatskar, M.; Yin, D.; Hsieh, C.-J.; and Chang, K.-W. 2019.
\newblock {VisualBERT}: A simple and performant baseline for vision and
  language.
\newblock \emph{arXiv}.

\bibitem[{Liu et~al.(2019)Liu, Huang, Zeng, Chen, and Fu}]{liu2019learning}
Liu, B.; Huang, Z.; Zeng, Z.; Chen, Z.; and Fu, J. 2019.
\newblock Learning Rich Image Region Representation for Visual Question
  Answering.
\newblock \emph{arXiv}.

\bibitem[{Lu et~al.(2019)Lu, Batra, Parikh, and Lee}]{lu2019vilbert}
Lu, J.; Batra, D.; Parikh, D.; and Lee, S. 2019.
\newblock Vilbert: Pretraining task-agnostic visiolinguistic representations
  for vision-and-language tasks.
\newblock In \emph{NeurIPS}.

\bibitem[{Lu et~al.(2020)Lu, Goswami, Rohrbach, Parikh, and Lee}]{Lu_2020_CVPR}
Lu, J.; Goswami, V.; Rohrbach, M.; Parikh, D.; and Lee, S. 2020.
\newblock 12-in-1: Multi-Task Vision and Language Representation Learning.
\newblock In \emph{CVPR}.

\bibitem[{Lu et~al.(2016)Lu, Yang, Batra, and Parikh}]{lu2016hierarchical}
Lu, J.; Yang, J.; Batra, D.; and Parikh, D. 2016.
\newblock {Hierarchical Question-Image Co-attention for Visual Question
  Answering}.
\newblock In \emph{NeurIPS}.

\bibitem[{Marino et~al.(2021)Marino, Chen, Parikh, Gupta, and
  Rohrbach}]{marino2020krisp}
Marino, K.; Chen, X.; Parikh, D.; Gupta, A.; and Rohrbach, M. 2021.
\newblock KRISP: Integrating Implicit and Symbolic Knowledge for Open-Domain
  Knowledge-Based VQA.
\newblock In \emph{CVPR}.

\bibitem[{Marino et~al.(2019)Marino, Rastegari, Farhadi, and
  Mottaghi}]{marino2019ok}
Marino, K.; Rastegari, M.; Farhadi, A.; and Mottaghi, R. 2019.
\newblock OK-VQA: A Visual Question Answering Benchmark Requiring External
  Knowledge.
\newblock In \emph{CVPR}.

\bibitem[{Narasimhan, Lazebnik, and Schwing(2018)}]{narasimhan2018out}
Narasimhan, M.; Lazebnik, S.; and Schwing, A. 2018.
\newblock Out-of-{T}he-{B}ox: {R}easoning with {G}raph {C}onvolution {N}ets for
  {F}actual {V}isual {Q}uestion {A}nswering.
\newblock In \emph{NeurIPS}.

\bibitem[{Narasimhan and Schwing(2018)}]{narasimhan2018straight}
Narasimhan, M.; and Schwing, A.~G. 2018.
\newblock Straight to the facts: Learning knowledge base retrieval for factual
  visual question answering.
\newblock In \emph{ECCV}.

\bibitem[{Qu et~al.(2021)Qu, Zamani, Yang, Croft, and
  Learned-Miller}]{dpr_okvqa}
Qu, C.; Zamani, H.; Yang, L.; Croft, W.~B.; and Learned-Miller, E. 2021.
\newblock Passage Retrieval for Outside-Knowledge Visual Question Answering.
\newblock In \emph{ACM SIGIR}.

\bibitem[{Raffel et~al.(2020)Raffel, Shazeer, Roberts, Lee, Narang, Matena,
  Zhou, Li, and Liu}]{2020t5}
Raffel, C.; Shazeer, N.; Roberts, A.; Lee, K.; Narang, S.; Matena, M.; Zhou,
  Y.; Li, W.; and Liu, P.~J. 2020.
\newblock Exploring the Limits of Transfer Learning with a Unified Text-to-Text
  Transformer.
\newblock \emph{JMLR}.

\bibitem[{Sharma et~al.(2018)Sharma, Ding, Goodman, and
  Soricut}]{sharma2018conceptual}
Sharma, P.; Ding, N.; Goodman, S.; and Soricut, R. 2018.
\newblock Conceptual captions: A cleaned, hypernymed, image alt-text dataset
  for automatic image captioning.
\newblock In \emph{ACL}.

\bibitem[{Shevchenko et~al.(2021)Shevchenko, Teney, Dick, and
  Hengel}]{shevchenko2021reasoning}
Shevchenko, V.; Teney, D.; Dick, A.; and Hengel, A. v.~d. 2021.
\newblock Reasoning over Vision and Language: Exploring the Benefits of
  Supplemental Knowledge.
\newblock \emph{arXiv}.

\bibitem[{Tan and Bansal(2019)}]{tan2019lxmert}
Tan, H.; and Bansal, M. 2019.
\newblock LXMERT: Learning Cross-Modality Encoder Representations from
  Transformers.
\newblock In \emph{EMNLP}.

\bibitem[{Tompson et~al.(2014)Tompson, Jain, LeCun, and
  Bregler}]{tompson2014joint}
Tompson, J.~J.; Jain, A.; LeCun, Y.; and Bregler, C. 2014.
\newblock Joint training of a convolutional network and a graphical model for
  human pose estimation.
\newblock In \emph{NeurIPS}.

\bibitem[{Turc et~al.(2019)Turc, Chang, Lee, and Toutanova}]{turc2019well}
Turc, I.; Chang, M.-W.; Lee, K.; and Toutanova, K. 2019.
\newblock Well-read students learn better: On the importance of pre-training
  compact models.
\newblock \emph{arXiv preprint arXiv:1908.08962}.

\bibitem[{Wang et~al.(2018)Wang, Wu, Shen, Dick, and van~den
  Hengel}]{wang2018fvqa}
Wang, P.; Wu, Q.; Shen, C.; Dick, A.; and van~den Hengel, A. 2018.
\newblock Fvqa: Fact-based visual question answering.
\newblock \emph{TPAMI}.

\bibitem[{Wang et~al.(2017)Wang, Wu, Shen, Hengel, and Dick}]{wang2015explicit}
Wang, P.; Wu, Q.; Shen, C.; Hengel, A. v.~d.; and Dick, A. 2017.
\newblock Explicit knowledge-based reasoning for visual question answering.
\newblock In \emph{IJCAI}.

\bibitem[{Wu, Chen, and Mooney(2020)}]{wu2020improving}
Wu, J.; Chen, L.; and Mooney, R.~J. 2020.
\newblock Improving VQA and its Explanations by Comparing Competing
  Explanations.
\newblock \emph{arXiv}.

\bibitem[{Wu et~al.(2016)Wu, Wang, Shen, Dick, and van~den
  Hengel}]{Wu_2016_CVPR}
Wu, Q.; Wang, P.; Shen, C.; Dick, A.; and van~den Hengel, A. 2016.
\newblock Ask Me Anything: Free-Form Visual Question Answering Based on
  Knowledge From External Sources.
\newblock In \emph{CVPR}.

\bibitem[{Yu et~al.(2019)Yu, Yu, Cui, Tao, and Tian}]{zhou2019deep}
Yu, Z.; Yu, J.; Cui, Y.; Tao, D.; and Tian, Q. 2019.
\newblock {Deep modular co-attention networks for visual question answering}.
\newblock In \emph{CVPR}.

\bibitem[{Zhang* et~al.(2020)Zhang*, Kishore*, Wu*, Weinberger, and
  Artzi}]{zhang2019bertscore}
Zhang*, T.; Kishore*, V.; Wu*, F.; Weinberger, K.~Q.; and Artzi, Y. 2020.
\newblock BERTScore: Evaluating Text Generation with BERT.
\newblock In \emph{ICLR}.

\bibitem[{Zhang et~al.(2020)Zhang, Deng, Ma, and
  Lam}]{zhang-etal-2020-answerfact}
Zhang, W.; Deng, Y.; Ma, J.; and Lam, W. 2020.
\newblock {A}nswer{F}act: Fact Checking in Product Question Answering.
\newblock In \emph{EMNLP}.

\bibitem[{Zhang, Vu, and Moschitti(2021)}]{zhang2021joint}
Zhang, Z.; Vu, T.; and Moschitti, A. 2021.
\newblock Joint Models for Answer Verification in Question Answering Systems.
\newblock \emph{arXiv}.

\bibitem[{Zhou et~al.(2020)Zhou, Palangi, Zhang, Hu, Corso, and
  Gao}]{zhou2020unified}
Zhou, L.; Palangi, H.; Zhang, L.; Hu, H.; Corso, J.~J.; and Gao, J. 2020.
\newblock Unified Vision-Language Pre-Training for Image Captioning and VQA.
\newblock In \emph{AAAI}.

\bibitem[{Zhu et~al.(2016)Zhu, Groth, Bernstein, and Fei-Fei}]{zhu2016visual7w}
Zhu, Y.; Groth, O.; Bernstein, M.; and Fei-Fei, L. 2016.
\newblock {Visual7w: Grounded Question Answering in Images}.
\newblock In \emph{CVPR}.

\bibitem[{Zhu et~al.(2020)Zhu, Yu, Wang, Sun, Hu, and Wu}]{zhu2020mucko}
Zhu, Z.; Yu, J.; Wang, Y.; Sun, Y.; Hu, Y.; and Wu, Q. 2020.
\newblock Mucko: Multi-Layer Cross-Modal Knowledge Reasoning for Fact-based
  Visual Question Answering.
\newblock In \emph{IJCAI}.

\end{thebibliography}

\end{document}